\documentclass[10pt,twocolumn,letterpaper]{article}

\usepackage{iccv}
\usepackage{times}
\usepackage{epsfig}
\usepackage{graphicx}
\usepackage{amsmath}
\usepackage{amssymb}

\usepackage{multirow}
\usepackage{enumitem}

\newcommand{\figref}[1]{Fig. \ref{#1}}
\newcommand{\tabref}[1]{Tab. \ref{#1}}

\renewcommand\footnotemark{}

\usepackage[breaklinks=true,colorlinks,bookmarks=false]{hyperref}

\iccvfinalcopy % *** Uncomment this line for the final submission

\def\iccvPaperID{8350} % *** Enter the ICCV Paper ID here

\ificcvfinal\fi

\begin{document}

%%%%%%%%% TITLE

\title{Learning Canonical 3D Object Representation for Fine-Grained Recognition\thanks{This work was supported by the National Research Foundation of Korea (NRF) grant funded by the Korea government (MSIP) (NRF-2021R1A2C2006703), and the Yonsei University Research Fund of 2021 (2021-22-0001).}}

\author{Sunghun Joung$^{1}$, Seungryong Kim$^{2}$, Minsu Kim$^{1}$, Ig-Jae Kim$^{3}$, Kwanghoon Sohn$^{1,*}$\thanks{$^{*}$Corresponding author}\\
$^1$Yonsei University,
$^2$Korea University,
$^3$Korea Institute of Science and Technology (KIST)\\
{\tt\small{\{sunghunjoung,minsukim320,khsohn\}@yonsei.ac.kr}},
\tt\small{seungryong\_kim@korea.ac.kr},
{\tt\small{drjay@kist.re.kr}}
}

% \author{First Author\\
% Institution1\\
% Institution1 address\\
% {\tt\small firstauthor@i1.org}
% % For a paper whose authors are all at the same institution,
% % omit the following lines up until the closing ``}''.
% % Additional authors and addresses can be added with ``\and'',
% % just like the second author.
% % To save space, use either the email address or home page, not both
% \and
% Second Author\\
% Institution2\\
% First line of institution2 address\\
% {\tt\small secondauthor@i2.org}
% }

\maketitle
% Remove page # from the first page of camera-ready.
\ificcvfinal\thispagestyle{empty}\fi

%%%%%%%%% ABSTRACT
\begin{abstract}
We propose a novel framework for fine-grained object recognition that learns to recover object variation in 3D space from a single image, trained on an image collection without using any ground-truth 3D annotation.
We accomplish this by representing an object as a composition of 3D shape and its appearance, while eliminating the effect of camera viewpoint, in a canonical configuration.
Unlike conventional methods modeling spatial variation in 2D images only, our method is capable of reconfiguring the appearance feature in a canonical 3D space, thus enabling the subsequent object classifier to be invariant under 3D geometric variation.
Our representation also allows us to go beyond existing methods, by incorporating 3D shape variation as an additional cue for object recognition.
To learn the model without ground-truth 3D annotation, we deploy a differentiable renderer in an analysis-by-synthesis framework.
By incorporating 3D shape and appearance jointly in a deep representation, our method learns the discriminative representation of the object and achieves competitive performance on fine-grained image recognition and vehicle re-identification.
We also demonstrate that the performance of 3D shape reconstruction is improved by learning fine-grained shape deformation in a boosting manner.
\end{abstract}

%%%%%%%%% BODY TEXT
\section{Introduction}\label{sec:1}
Object recognition \cite{krizhevsky2012imagenet,he2016deep,fu2017look,zheng2019looking} is one of the most fundamental and essential tasks in computer vision fields, which has achieved steady progress by the advent of deep convolutional neural networks.
However, it still remains a challenging problem, especially when an object undergoes severe geometric deformations, \eg, by object scale, pose and part variations, which frequently occur across different instances, or by camera viewpoint changes~\cite{felzenszwalb2009object,jaderberg2015spatial,dai2017deformable,joung2020cylindrical}.

\begin{figure}[!t]
\centering
{\includegraphics[width=1\linewidth]{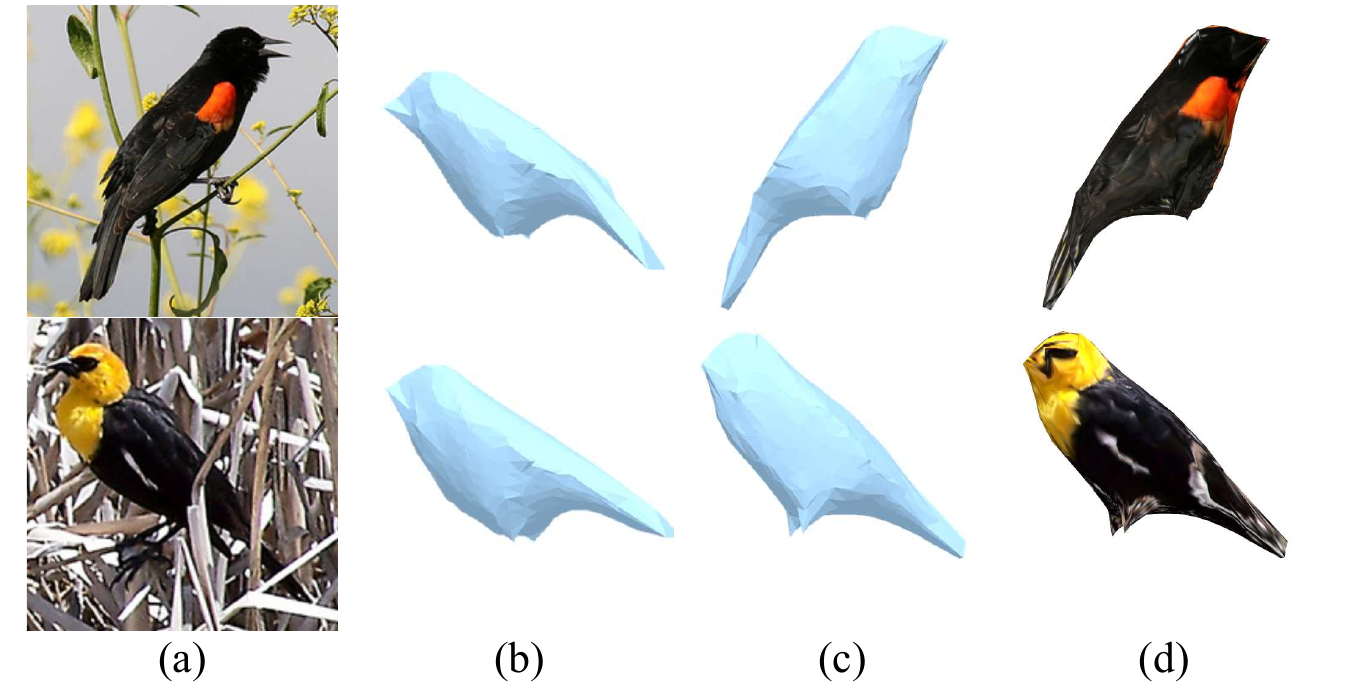}}\\\vspace{-2pt}
\caption{
\textbf{Intuition of our method:} Given (a) 2D image, we recover object variation in 3D space by estimating (b) 3D shape deformation, (c) camera viewpoint change and (d) appearance variation.
It allows for using 3D shape and appearance in a canonical space, while eliminating camera viewpoint variation, enabling us to deal with 3D object variations and facilitating the subsequent object classifier.}
\label{fig:1}
\vspace{-10pt}
\end{figure}

To overcome these challenges, recent works~\cite{jaderberg2015spatial,lin2017inverse,shu2018deforming,dai2016r,dai2017deformable} seek to handle such geometric variations based on an assumption that the object variation can be decomposed into \emph{appearance} and \emph{2D spatial variation}.
They first estimate an appearance flow from an input and then warp the input into a canonical configuration so as to remove the spatial variation, from which the appearance feature is extracted to facilitate the subsequent classifier's task. The appearance flow is generally estimated by modeling  2D transformation~\cite{jaderberg2015spatial,lin2017inverse,shu2018deforming}, \eg, affine transformation, or by learning offset of sampling locations in the convolutional operators~\cite{dai2016r,dai2017deformable}.
These methods, however, do not account for the fact that the object variation, given an image, is due to the variations in \emph{appearance}, \emph{3D shape} and \emph{camera viewpoint} as in \figref{fig:1}.
While the effect of camera viewpoint should be eliminated for achieving geometric invariance, 3D shape variation can be used as an additional cue to extract a shape feature that is able to supplement an appearance feature, but none of the existing methods utilize this.

However, estimating 3D object information from a single image is challenging, since collecting the ground-truth 3D shape is notoriously difficult and time-consuming~\cite{xiang2014beyond}, thus limiting the supervised learning for this task.
To overcome this, some methods implicitly consider the 3D object structure by learning a discriminative feature representation for fine-grained recognition.
Formally, they use an extra module to localize the discriminative object parts \cite{krause20133d,guo2019aligned}, by using explicit part detectors \cite{branson2014bird,krause2015fine} or implicit attention mechanisms \cite{fu2017look,zheng2017learning}.
However, these methods can only localize a few semantic parts without understanding the holistic object structure, and can be limited if the network fails to consistently localize object parts across multiple instances.

In this paper, we present a method that estimates 3D object information in a canonical configuration, including 3D object shape and appearance, with camera viewpoint.
It enables the subsequent classifier to directly work on the 3D object information, from which both appearance and shape features are simultaneously extracted.
It allows for handling subtle intra-class variations by means of both appearance and 3D shape features, which is not available at the existing approaches.
To this end, we deploy a differentiable renderer \cite{kato2018neural,liu2019soft}, to infer 3D shape, without ground-truth 3D annotation, in an analysis-by-synthesis framework, as in recent 3D shape reconstruction methods \cite{kanazawa2018learning,insafutdinov2018unsupervised,tatarchenko2019single,navaneet2020image}.
In particular, we incorporate this framework into an encoder-decoder architecture that disentangles the object variation to 3D shape, appearance, and camera viewpoint.
To this end, an image is embedded into a low-dimensional latent code that is fed into separate decoders to estimate the aforementioned factors independently.
We also exploit multiple hypothesis camera prediction to avoid local minima during training as in \cite{insafutdinov2018unsupervised,kulkarni2019canonical,goel2020shape}.

Unlike conventional methods \cite{jaderberg2015spatial,lin2017inverse,shu2018deforming} that model 2D spatial variation only, our method is capable of reconfiguring an appearance feature in a canonical space, where each semantic part of an object is mapped to the same location in a canonical space.
Moreover, our method enables dense semantic alignment \cite{zhang2019densely} into a canonical configuration, where positional encoding \cite{liu2018intriguing,carion2020end} can further improve recognition performance, while the conventional methods \cite{branson2014bird,krause2015fine,fu2017look,zheng2017learning} are limited by highlighting only a few salient parts of an object.
To improve recognition ability between subtle object variations, we further introduce a shape encoder to utilize 3D shape deformation as an additional cue.
By incorporating 3D shape and appearance jointly in a deep representation, our method consistently boosts the discriminative representation learning on fine-grained image recognition and vehicle re-identification tasks.
In addition, our joint learning framework enables us to improve 3D shape reconstruction capability by discriminating shape variations between different fine-grained categories.

%-------------------------------------------------------------------------

\section{Related Works}\label{sec:2}

\paragraph{Spatial Invariance.}\label{sec:21}
Vanilla CNN \cite{simonyan2014very,he2016deep} provided limited performance under severe geometric variations.
STN \cite{jaderberg2015spatial} offered a way to explicitly handle geometric variation by spatially warping the input to a canonical configuration to facilitate the recognition task.
Inspired by STN \cite{jaderberg2015spatial}, many variants were proposed by using recurrent formalism \cite{lin2017inverse}, deformable convolution \cite{dai2016r}, polar transformation \cite{esteves2018polar}, and attention based warping \cite{recasens2018learning,zheng2019looking}.
These methods typically employ an additional localization network, to predict appearance flow, which is then applied to intermediate features to remove spatial variation.
Since they do not share the template shape with different images, modeling shape variability within a category is limited.
On the other hand, several methods attempted to address the geometric variation of an object by decomposing the image into shape and appearance with instance-agnostic template shape.
Thewlis \etal exploited dense coordinate frame \cite{thewlis2017unsupervised,thewlis2018modelling} in order to recover the deformation of an object with equivariance.
Deforming autoencoders \cite{shu2018deforming} proposed a generative model to predict a deformation field by disentangling shape and appearance.
While effective, all of these methods model geometric deformation in 2D space, so they lack robustness against geometric variation in 3D space including 3D shape deformation and camera viewpoint variation.

\vspace{-10pt}
\paragraph{Fine-grained Object Recognition.}\label{sec:22}
Since modeling severe deformations of 3D object in 2D image is challenging, conventional methods for fine-grained object recognition \cite{farrell2011birdlets,krause20133d,branson2014bird,krause2015fine,lin2015bilinear} aim to learn discriminative feature representation of object parts and then classify the object based on the discriminative regions.
This, however, requires large human efforts as it needs extra annotation of bounding boxes or parts.
To alleviate this, recent methods proposed to automatically localize the discriminative object parts without part annotation using attention mechanisms \cite{fu2017look,zheng2017learning,sun2018multi,zheng2019looking,kim2020volumetric} in an unsupervised manner.
However, a deeper understanding of the holistic 3D object shape is essential as the network may not consistently localize the discriminative object parts across multiple instances.
Another line of research focuses on end-to-end feature encoding \cite{ge2019weakly,luo2019cross,ding2019selective,du2020fine,zhao2021graph} to encourage feature discriminability using deeper representations, high-order feature interactions or metric learning.

\vspace{-10pt}
\paragraph{Vehicle Re-identification.}\label{sec:23}
Vehicle re-identification task has gained more attention in recent years, following the release of several benchmarks \cite{liu2016deep,liu2016deep2}.
The main focus of the task has been on addressing viewpoint variation from 2D images.
Given vehicle images under arbitrary camera viewpoints, recent methods aim to transform the input appearance feature into a viewpoint independent representation.
Therefore, they utilize either local region based feature learning on pre-defined distinctive regions or keypoints \cite{wang2017orientation,he2019part,meng2020parsing,chen2020orientation}, or attention mechanisms \cite{zhou2018view,khorramshahi2019dual}.
\begin{figure*}[!t]
\centering
{\includegraphics[width=1\linewidth]{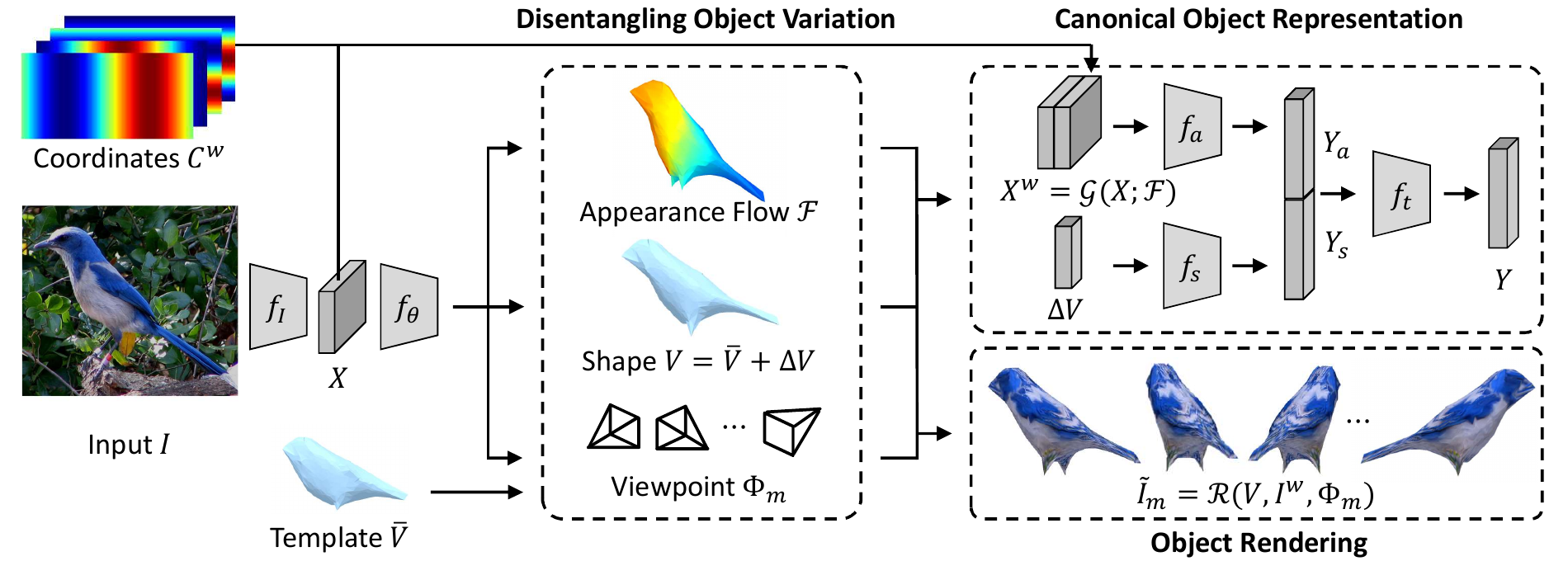}}\\
\vspace{-2pt}
\caption{
\textbf{Overview of our framework:}
The input feature $X$, from an image encoder ${f}_{I}$, is fed into a module ${f}_{\theta}$ for disentangling object variation into appearance flow $\mathcal{F}$, shape deformation $\Delta{V}$ and camera viewpoint ${\Phi}$ as well as parameterized template shape $\overline{V}$, respectively.
Then appearance feature ${Y}_{a}$ and shape feature ${Y}_{s}$ are obtained by applying the appearance encoder ${f}_{a}$ and shape encoder ${f}_{s}$ to get the final object representation $Y$.
In our framework, the entire network is trained in a joint and boosting manner through fine-grained object recognition and 3D shape reconstruction tasks.
}
\label{fig:2}\vspace{-10pt}
\end{figure*}

\vspace{-10pt}
\paragraph{Single-view 3D Reconstruction.}\label{sec:24}
Single-view 3D reconstruction methods aim to reconstruct 3D object shape from a single image.
While conventional methods utilize ground-truth 3D shape for training \cite{wang2018pixel2mesh,wen2019pixel2mesh++}, it requires tremendous human efforts for annotation \cite{xiang2016objectnet3d}, or their applicability is restricted to synthetic data \cite{chang2015shapenet}.
Therefore, several methods have proposed a differentiable renderer \cite{kato2018neural,liu2019soft} to train 3D reconstruction networks using either multi-view images or ground-truth camera viewpoints in an analysis by synthesis framework.
To relax such constraints on supervision, Kanazawa \etal proposed CMR \cite{kanazawa2018learning} to explore 3D reconstruction from a collection of images with different instances, by exploiting a learnable 3D template shape.
Since CMR \cite{kanazawa2018learning} requires annotated 2D keypoints for training, several works were proposed to mitigate this by using camera-multiplex \cite{goel2020shape}, semantic consistency \cite{li2020self} or temporal consistency \cite{li2020online}, which can estimate 3D mesh with foreground mask for training.

%-------------------------------------------------------------------------

\section{Method}\label{sec:3}
\subsection{Preliminaries}\label{sec:31}
Let us denote an intermediate CNN feature representation as $X\in{\mathbb{R}}^{H\times W\times K}$, with height $H$, width $W$ and $K$ channels.
To reduce the spatial variations among different instances within the representation, recent works \cite{jaderberg2015spatial,lin2017inverse} predict an appearance flow $\mathcal{F}$ to transform the input feature into a canonical configuration, producing a warped feature ${X}^w=\mathcal{G}(X;\mathcal{F})$ through sampling function $\mathcal{G}$, \eg, a bilinear sampler \cite{jaderberg2015spatial}.
The appearance flow is estimated via a module ${f}_{\theta}$, that takes $X$ as an input and outputs the transformation parameter (\eg, affine transformation) to produce appearance flow $\mathcal{F}\in{\mathbb{R}}^{{H}^w\times{W}^w\times2}$, where each value in $\mathcal{F}(u,v)$ at point $(u,v)$ indicates the coordinates of the input to be sampled, and ${H}^w$, ${W}^w$ are the height and width of ${X}^w$.
Conventional methods solely consider 2D geometric deformation and extract the features for appearance only \cite{jaderberg2015spatial,dai2017deformable}.
However, since the geometric variation of objects occurs in 3D space, the warped feature is often inconsistent across instances under different camera viewpoints.

\subsection{Motivation and Overview}\label{sec:32}
In this paper, we conjecture that the object variation can be further decomposed into variations of 3D shape and appearance as well as camera viewpoint changes, and by only removing the latter, we can effectively handle complex object variation for fine-grained recognition.
In other words, both 3D shape and appearance variation have to be encoded into object representation to discriminate subtle intra-class variation.
The main bottleneck is learning to reconstruct the 3D shape.
Traditional methods for 3D shape estimation, in training, relied on datasets labeled with 3D shapes \cite{choy20163d,fan2017point} or multiple views of the same object \cite{tulsiani2017multi,insafutdinov2018unsupervised}.
Recent works \cite{kanazawa2018learning,goel2020shape,li2020self} have relaxed the constraints of supervision by utilizing an analysis-by-synthesis framework that renders an instance-agnostic template shape into 2D images via a differentiable renderer \cite{kato2018neural,liu2019soft}.

Inspired by this, we present a framework that learns the mapping function ${f}_{\theta}$ to recover 3D object information from a single image as illustrated in \figref{fig:2}.
By disentangling object variation into a composition of independent factors of 3D shape, appearance, and camera viewpoint variation, each component can be trained in a joint and boosting manner through object recognition and 3D shape reconstruction tasks.
In order to learn these mappings without ground-truth 3D annotation, we represent the template shape of an object as 3D mesh in a canonical space, where the predicted object variation projects the mesh from this canonical space to the image coordinates.
The use of canonical 3D space allows for assigning dense semantic correspondences across different instances to be consistent.
At the same time, we estimate an appearance flow to transform the 2D image into the canonical space, where the warped feature ${X}^{w}$ to be spatially invariant under 3D geometric variation.
Since shape deformation itself can be an additional cue to discriminate object classes, we further present the shape encoder to maximize the classification performance.

\subsection{Disentangling Object Variation}\label{sec:33}
We model ${f}_{\theta}$ to project the input feature $X$ into a low dimensional latent code that is fed into three decoders to predict an appearance flow $\mathcal{F}$, shape deformation $\Delta{V}$, and camera viewpoint $\Phi$, such that $\{\mathcal{F},\Delta{V},\Phi\}={f}_{\theta}(X)$.

\vspace{-10pt}
\paragraph{Appearance flow $\mathcal{F}$.}\label{sec:331}
We first learn an appearance flow $\mathcal{F}$ to transform the input image defined in 2D to the canonical configuration so as to remove 3D geometric variation and enable to align appearance feature with similar semantics to the same location in a canonical space.
As the topology of 3D mesh is fixed, we use ${I}^{w}=\mathcal{G}(I;\mathcal{F})$ to model the texture of the template mesh for object rendering as in~\cite{kanazawa2018learning,li2020self}.
For appearance representation in a canonical space, we use ${X}^{w}=\mathcal{G}(X;\mathcal{F})$, where we take advantage of a pre-trained network, \ie, ResNet-50~\cite{he2016deep}.

\vspace{-10pt}
\paragraph{Shape deformation $\Delta{V}$.}\label{sec:332}
\begin{figure}[!t]
\centering
{\includegraphics[width=1\linewidth]{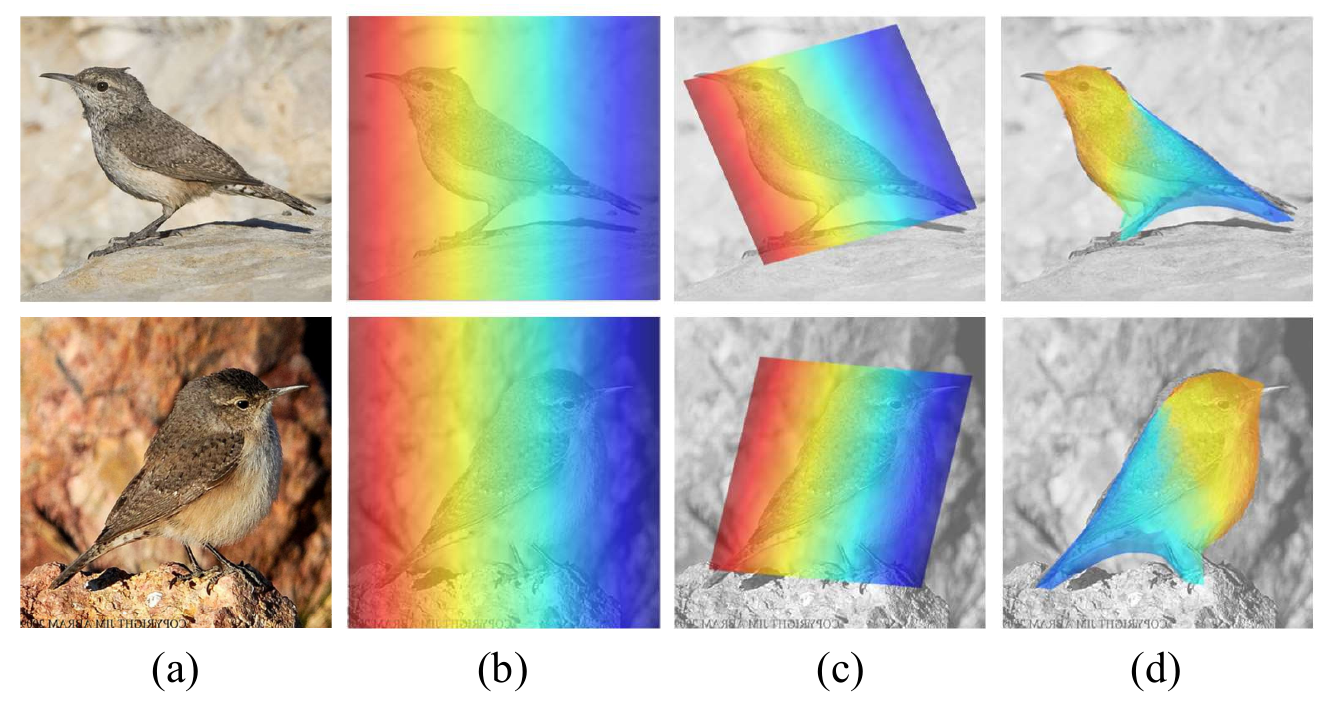}}\\
\vspace{-2pt}
\caption{
\textbf{Comparison of appearance flow:} (a) input images, (b) source coordinates in an image, and appearance flow obtained using (c) STN~\cite{jaderberg2015spatial}, and (d) ours.
Points with the same color in different images are projected to the same point in a canonical space.
This shows that our method can make the warped feature to be densely consistent across different camera viewpoints, while the existing methods fail (Best viewed in color).}
\vspace{-10pt}
\label{fig:3}
\end{figure}
We represent the 3D shape in a form of mesh $M=(V,F)$ with vertices $V\in{\mathbb{R}}^{\vert{V}\vert\times3}$ and faces $F$.
The set of faces defines the connectivity of vertices in the spherical mesh, and we assume it remains fixed.
The vertex positions of a deformable object are determined as the summation of an instance-specific deformation $\Delta{V}$ predicted from an image to a learned instance-agnostic mean shape $\overline{V}$ such that $V=\Delta{V}+\overline{V}$.
Since most object categories exhibit bilateral symmetry \cite{wu2020unsupervised,joung2020cylindrical}, we further constrain the predicted shape and deformation to be mirror-symmetric, following \cite{kanazawa2018learning}.
This symmetric constraint can be utilized to reduce the number of parameters for shape representation as well as infer the invisible surface of an object from the visible features. 
We leverage $\Delta{V}$ to extract shape variation features. 

\vspace{-10pt}
\paragraph{Camera viewpoint $\Phi$.}\label{sec:333}
For camera parameters, we assume a weak-perspective projection, parameterized by scale $\mathbf{s}\in\mathbb{R}$, translation $\mathbf{t}\in{\mathbb{R}}^{2}$, and rotation captured by quaternion $\mathbf{r}\in{\mathbb{R}}^{4}$.
We use $\Phi=(\mathbf{s},\mathbf{t},\mathbf{r})$ to denote the projection of 3D points sets on template shape into 2D image coordinates via the weak perspective projection.
We optimize the overall network over the multi-hypothesis camera viewpoint~\cite{insafutdinov2018unsupervised,kulkarni2019canonical,goel2020shape}, which has been well-known to be robust, by maintaining a set of possible camera hypotheses for each training instance, where $\mathcal{C}=\{{\Phi}_{1},...,{\Phi}_{M}\}$ denotes viewpoints with $M$ cameras.

\subsection{Canonical Object Representation}\label{sec:34}

In this section, we introduce canonical object representation that exploits the feature from appearance and shape deformation, invariant under arbitrary camera viewpoints.

\vspace{-10pt}
\paragraph{Embedding appearance.}\label{sec:341}
As argued above, we explicitly learn to map the pixels in the object to their corresponding locations on template shape.
It makes each point in ${X}^{w}$ densely and semantically aligned to a canonical 3D space as shown in \figref{fig:3}.
Note that conventional methods cannot enforce the warped feature to be consistent under different camera viewpoint \cite{jaderberg2015spatial,lin2017inverse,esteves2018polar}, or only can localize a few semantic parts \cite{branson2014bird,krause2015fine,fu2017look,zheng2017learning}.

In addition, to take advantage of the positional sensitivity, where semantically similar parts are densely mapped to a canonical space, we utilize positional encoding (PE), which has been shown to be effective in natural language processing \cite{vaswani2017attention,devlin2019bert} and object recognition \cite{liu2018intriguing,carion2020end}.
The straightforward way is to exploit a coordinate map of the same spatial dimensions as ${X}^{w}$, normalized to be $u,v\in[-1,1]$, following CoordConv \cite{liu2018intriguing}.
This, however, cannot model the continuity of ${X}^{w}$ on a 3D mesh surface that is not in 2D planar space.
To overcome this, we modify the original 2-dimensional $(u,v)$ pixel coordinates into 4-dimensional coordinates $(\mathrm{cos}(\pi u),\mathrm{sin}(\pi u),\mathrm{cos}(\pi v),\mathrm{sin}(\pi v))$, to provide periodic continuity of the 2D coordinates on 3D the surface.
This canonical position map ${C}^{w}$ is then concatenated to the input ${X}^{w}$ and passed to the appearance encoder ${f}_{a}$ with two convolutional layers to output appearance representation ${Y}_{a}$ in a vectorized form.
With this simple technique, the network can disambiguate different positions of an object instance, and also model the structural composition of object parts.

\vspace{-10pt}
\paragraph{Embedding shape deformation.}\label{sec:342}
\begin{figure*}
\centering
{\includegraphics[width=1\linewidth]{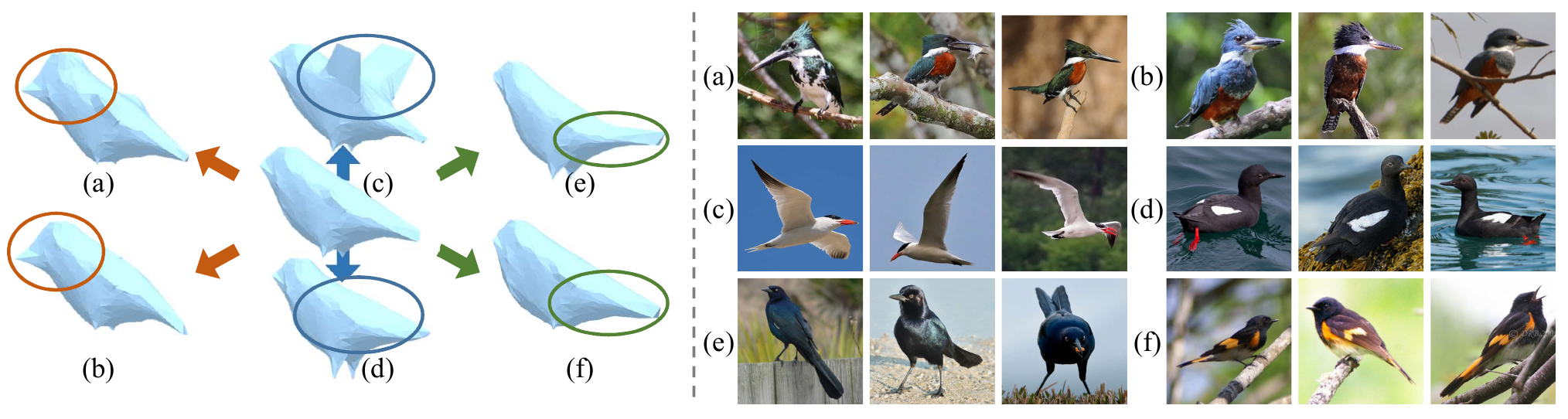}}\\
\vspace{-2pt}
\caption{\textbf{Visualization of the learned 3D object shape deformations:} We visualize a template shape $\overline{V}$ and averaged shapes of 6 different fine-grained categories.
Each shape characterizes a fine-grained category of, for instance, birds on (a,b) head, (c,d) body, and (e,f) tail types.
We utilize such shape deformation as an additional cue to discriminate subtle intra-class variation for the object recognition task.
}
\label{fig:4}\vspace{-10pt}
\end{figure*}
Unlike existing methods \cite{jaderberg2015spatial,lin2017inverse,esteves2018polar} that estimate geometric variation to remove spatial variation, we take advantage of modeling 3D shape.
Note that the intra-class shape variation of a particular category has been utilized for 3D shape recognition \cite{xiang2014beyond,chang2015shapenet} using 3D model as input, but none of the existing methods utilize it for 2D images.
Rather than directly encoding shape representation $V$, we exploit instance-specific shape deformation $\Delta{V}$ since it shares the learned instance-agnostic mean shape $\overline{V}$, which is more suitable to discriminate subtle differences.
We thus present an additional shape encoder, where the vectorized $\Delta{V}$ is fed into the shape encoder ${f}_{s}$ with several fully-connected layers to output shape representation ${Y}_{s}$.

\vspace{-10pt}
\paragraph{Fusing appearance and shape representation.}\label{sec:343}
Given feature representation of appearance ${Y}_{a}$ and shape ${Y}_{s}$ in a canonical space, we concatenate them together to combine appearance and shape variations and passed to the target network ${f}_{t}$ with two fully-connected layers for the final object representation $Y$.

\subsection{Loss Function}\label{sec:35}

Since we train the networks on an image collection without any ground-truth 3D shape, multi-view, camera viewpoints or keypoint supervision, we follow unsupervised learning of category-specific mesh reconstruction by utilizing analysis-by-synthesis framework~\cite{goel2020shape}.
We predict multiple camera hypotheses to overcome a local minima, by making every instance maintain its own $\mathcal{C}$ independently, where we compute the loss against each camera viewpoint ${\Phi}_{m}$.
By jointly learning our networks on fine-grained object classification and 3D shape reconstruction tasks, appearance flow $\mathcal{F}$ and shape deformation $\Delta{V}$ can be trained in a way that mutually boosts the two tasks.
In the following, we describe loss functions to train our network in detail.

\vspace{-10pt}
\paragraph{ Loss for disentangling.}\label{sec:351}
We denote the rendered image as ${\tilde{I}}_{m}=\mathcal{R}(V,{I}^{w},{\Phi}_{m})$ and rendered silhouette mask as ${\tilde{S}}_{m}=\mathcal{R}(V,{\Phi}_{m})$.
We then compute the silhouette loss ${\mathcal{L}}_{\mathrm{mask},m}$ and image reconstruction loss ${\mathcal{L}}_{\mathrm{pixel},m}$ for $m$-th camera viewpoint as follows:
\begin{equation}\label{equ:loss_mask}
\mathcal{L}_{\mathrm{mask},m}={\Vert S-{\tilde{S}}_{m}\Vert}_{2}^{2}+\mathrm{dt}(S)*{\tilde{S}}_{m},
\end{equation}
\begin{equation}\label{equ:loss_pixel}
\mathcal{L}_{\mathrm{pixel},m}=\mathrm{dist}(\tilde{I}_{m}\odot S, I\odot S),
\end{equation}
where $\mathrm{dt}(\cdot)$ denotes distance transform, $*$ is matrix multiplication, $\odot$ is element-wise multiplication, and $\mathrm{dist}(\cdot)$ represents a perceptual distance metric \cite{zhang2018unreasonable}.
Note that we only use foreground mask $S$ for training.

\vspace{-10pt}
\paragraph{ Loss for priors.}\label{sec:352}
In order to recover 3D shape of smooth surface, we apply smoothness loss ${\mathcal{L}}_{\mathrm{smooth}}={\Vert LV\Vert}_{2}$, where $L$ is the discrete Laplace-Beltrami operator to minimize the mean curvature \cite{pinkall1993computing}.
We construct $L$ once using the initial template mesh following our baseline \cite{kanazawa2018learning}.
Furthermore, we apply deformation regularization loss ${\mathcal{L}}_{\mathrm{reg}}={\Vert{\Delta{V}}\Vert}_{2}$ to prevent large deformations.

\vspace{-10pt}
\paragraph{ Overall training objective.}\label{sec:353}
Since we predict a set of multiple hypothesis for camera projection, we first define the summation of silhouette and image reconstruction loss ${\mathcal{L}}_{m}={\mathcal{L}}_{\mathrm{mask},m}+{\mathcal{L}}_{\mathrm{pixel},m}$ over the losses of $M$ camera pose prediction.
We then compute ${p}_{m} = \frac{{e}^{-{L}_{m}/\sigma}}{\sum_{n}{{e}^{-{L}_{n}/\sigma}}}$, the probability of being the optimal camera, to associate with each ${L}_{m}$ with its probability ${p}_{m}$.
The overall training objective is as follows, where we normalize each energy term according to its magnitudes as in \cite{kanazawa2018learning} with task-specific loss function ${\mathcal{L}}_{\mathrm{task}}$:
\begin{equation}\label{equ:loss_total}
{\mathcal{L}}_{\mathrm{total}}=
\sum\nolimits_{m}{p}_{m}{\mathcal{L}}_{\mathrm{m}}+{\mathcal{L}}_{\mathrm{smooth}}+{\mathcal{L}}_{\mathrm{reg}}
+{\mathcal{L}}_{\mathrm{task}}.
\end{equation}

\subsection{Implementation Details}\label{sec:36}
We implement our method using the Pytorch library \cite{paszke2017automatic}.
In our experiments, we utilize ResNet-50 \cite{he2016deep} pretrained on ImageNet \cite{deng2009imagenet} as backbone.
We build our module ${f}_{\theta}$ on the last convolutional layers of ResNet-50 \cite{he2016deep}.
For mesh representation, we use 642 vertices and 1280 faces correspond to Icosphere.
We exploit 305 symmetric vertex pairs and 32 vertices without symmetry, resulting in $\vert V\vert=337$.
For the encoder, we follow \cite{kanazawa2018learning}, by first applying a convolutional layer to downsample the spatial and channel dimensions into 1/4 and 1/8, respectively, which is then vectorized to form a 4,096-D vector.
We then apply two fully-connected layers to get the shared latent code of size 200.
This latent code is then fed into independent decoder networks with linear layers to predict shape deformation ${\mathbb{R}}^{\vert{V}\vert\times3}$ and camera projection parameters ${\mathbb{R}}^{7}$.
For the appearance flow, the latent code is fed into 5 upconvolution layers followed by $\mathrm{tanh}$ function to normalize the output into the $[-1,1]$ coordinate space.

In addition, the input images are resized to a fixed resolution of $512\times512$, and ${H}^{w},{W}^{w}$ are set as $256,512$.
Since the spatial resolution is different between $I$ and $X$, we set ${H}^{w},{W}^{w}$ to learn texture mapping ${I}^{w}$, and downsample it into $1/32$ for warping feature $X$ into ${X}^{w}$.
For task-specific loss function ${\mathcal{L}}_{\mathrm{task}}$, we use the cross-entropy loss for fine-grained image recognition, and use cross-entropy and triplet loss for vehicle re-identification as in \cite{meng2020parsing}.
We use 3D template meshes in \cite{kulkarni2019canonical} as initial meshes of birds and cars, since it speeds up to convergence of the networks \cite{kanazawa2018learning,goel2020shape}.
For the mask label, we use ground-truth mask on CUB-Birds \cite{wah2011caltech}, and obtain fore-ground masks using off-the-shelf segmentation \cite{he2017mask} for Stanford-Cars \cite{krause20133d} and Veri-776 \cite{liu2016deep}.
For all experiments, we use SoftRasterizer \cite{liu2019soft} as our 3D mesh rendering module $\mathcal{R}$ and follow the experimental protocols of \cite{goel2020shape} for training.

%-------------------------------------------------------------------------

\section{Experiments}\label{sec:4}
\subsection{Experimental Setup}\label{sec:41}
In this section, we comprehensively analyze and evaluate our method on fine-grained image recognition and vehicle re-identification.
First, we analyze the influence of the different components of our method on fine-grained image recognition.
We then evaluate our method compared to the state-of-the-art methods.
Finally, we evaluate 3D shape reconstruction performance compared to existing methods.

\begin{table}[!t]
    \centering
    \begin{tabular}{ccc}
    \hline
    Methods & CUB-Birds \cite{wah2011caltech} & Stanford-Cars \cite{krause20133d} \\ \hline \hline
    Base \cite{he2016deep} & 74.6 & 70.4 \\ \hline
    STN \cite{jaderberg2015spatial} & 76.5 & 71.0 \\
    DCN \cite{dai2017deformable} & 76.7 & 72.1 \\
    SSN \cite{recasens2018learning} & 77.7 & 74.8 \\
    ASN \cite{zheng2019looking} & 78.9 & 75.2 \\
    VTN \cite{kim2020volumetric} & 83.1 & 82.7 \\ \hline
    Ours w/o PE, ${f}_{s}$  & 83.7 & 85.5 \\
    Ours w/o ${f}_{s}$  & 86.8 & 93.2 \\
    Ours  & \bf{88.4} & \bf{94.7} \\ \hline
    \end{tabular}
    \vspace{6pt}
    \caption{Ablation study for the different components of our method on fine-grained image recognition.
    }\label{tab:1}
    \vspace{-10pt}
\end{table}

\subsection{Ablation Study}\label{sec:42}
We first analyze our method with the ablation studies, with respect to the different components, different PE modules and the different number of layers in the shape encoder, followed by visual analysis of shape deformation.

\vspace{-10pt}
\paragraph{Analysis of the different components.}\label{sec:421}
To validate the geometric invariance of our method, we compare with previous spatial deformation modeling methods, such as STN \cite{jaderberg2015spatial}, DCN \cite{dai2017deformable}, SSN \cite{recasens2018learning}, ASN \cite{zheng2019looking} and VTN \cite{kim2020volumetric} on fine-grained image recognition benchmarks including CUB-Birds \cite{wah2011caltech} and Stanford-Cars \cite{krause20133d}.
For a fair comparison, we apply these methods at the same layers as ours, \ie, the last convolutional layers of ResNet-50 \cite{he2016deep}.
As an ablation study, we evaluate our method with different components, only with spatial deformation modeling without positional encoding and shape encoder, denoted by Ours w/o PE, ${f}_{s}$ and without shape encoder, denoted by Ours w/o ${f}_{s}$.
The results are provided in \tabref{tab:1}, where our method consistently outperforms the conventional methods.
It is noticeable that the positional encoding module can only be used in our method since conventional methods \cite{jaderberg2015spatial,recasens2018learning,zheng2019looking} cannot map pixels on the same semantic part to a canonical space across different camera viewpoints.
Furthermore, our method can use the holistic 3D structure of an instance thanks to its invariant nature under 3D geometric variation including 3D object shape and camera viewpoint variation.

\vspace{-10pt}
\paragraph{The effects on different PE module.}\label{sec:422}
\begin{figure*}[!t]
\centering
{\includegraphics[width=1\linewidth]{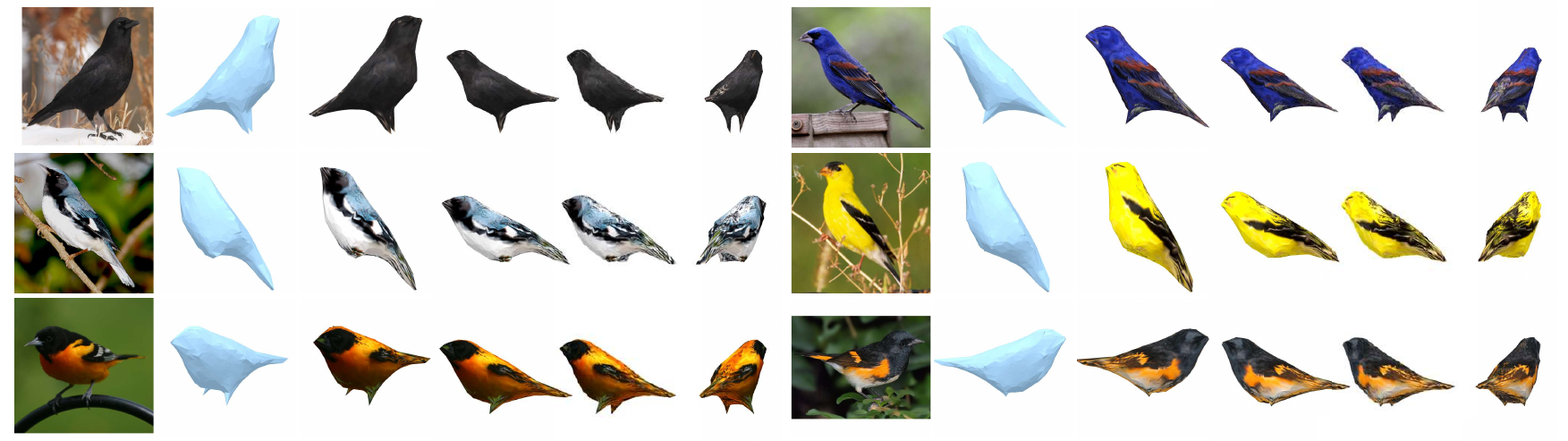}}\\
\vspace{-2pt}
\caption{\textbf{Qualitative examples on CUB-Birds \cite{wah2011caltech}:} (from left) input image $I$, its 3D shape $V$, and its rendered results ${\tilde{I}}_{m}$ from predicted camera viewpoint and from three other camera viewpoints.
}
\label{fig:5}\vspace{-10pt}
\end{figure*}
\begin{table}[!t]
    \centering
    \begin{tabular}{ccc}
    \hline
    PE Module & CUB-Birds \cite{wah2011caltech} & Stanford-Cars \cite{krause20133d} \\ \hline \hline
    None & 83.7 & 85.5 \\ \hline
    PE-2  & 85.2 & 89.8 \\
    PE-4  & \bf{86.8} & \bf{93.2} \\ \hline
    \end{tabular}
    \vspace{6pt}
    \caption{Ablation study for the different positional encoding (PE) modules on fine-grained image recognition.
    }\label{tab:2}
\end{table}
\begin{table}[!t]
    \centering
    \begin{tabular}{ccc}
    \hline
    \# of layers & CUB-Birds \cite{wah2011caltech} & Stanford-Cars \cite{krause20133d} \\ \hline \hline
    0 & 86.8 & 93.2 \\ \hline
    1 & 87.0 & 93.6 \\
    2 & 87.6 & 94.3 \\
    3 & \bf{88.4} & \bf{94.7} \\
    4 & 87.9 & 94.4 \\ \hline
    \end{tabular}
    \vspace{6pt}
    \caption{Ablation study for the different number of fully connected layers in shape encoder ${f}_{s}$ on fine-grained image recognition.
    }\label{tab:3}
    \vspace{-10pt}
\end{table}
\tabref{tab:2} shows experiments to validate the effect of various PE modules.
We denote the method using 2-dimensional $u,v$ pixel coordinates \cite{liu2018intriguing} as PE-2, and the proposed methods using 4-dimensional coordinates as PE-4.
Both PE-2 and PE-4 have shown higher accuracy by favoring a position sensitivity in a canonical appearance space.
These results share the same properties as previous studies \cite{liu2018intriguing,carion2020end}, indicating that positional encoding benefits to clarify the spatial representation.
In addition, PE-4 has shown better performance by utilizing sinusoidal functions to provide periodic continuity on 3D surface, rather than 2D space as in PE-2.

\vspace{-10pt}
\paragraph{The effects on different shape encoder.}\label{sec:423}

Since there is no reference model to utilize shape deformation from 2D images, we evaluate the performance with respect to the different number of layers in the model.
For simplicity, we fixed the channel dimension of each fully connected layer to be 512, followed by ReLU, with a different number of layers.
As the result with 3 fully connected layers has shown the best performance in \tabref{tab:3}, we set the number of layers as 3 for the remaining experiments.

\vspace{-10pt}
\paragraph{Visual analysis of shape deformation.}\label{sec:424}
To analyze the discriminate capability of shape feature, we visualize the learned shape deformations in the validation set of CUB-Birds \cite{wah2011caltech} as exemplified in \figref{fig:4}.
We averaged the estimated $\Delta{V}$ for each fine-grained category, which is associated with learned mean shape $\overline{V}$ such that $V=\overline{V}+\Delta{V}$ for visualization.
We can see that each shape corresponds to the natural factors of fine-grained categories, such as long beak or round tail, and the statistics of captured scenes, such as a bird flying in the sky or floating on the water.

\begin{table}[!t]
    \centering
    \begin{tabular}{cccc}
    \hline
    Methods & Backbone & \cite{wah2011caltech} & \cite{krause20133d} \\ \hline \hline
    PN-CNN \cite{branson2014bird} & AlexNet & 85.4 & - \\
    SPDA-CNN \cite{zhang2016spda} & VGG-19  & 85.1 & - \\ \hline
    PA-CNN \cite{krause2015fine} & VGG-19 & 82.8 & 92.8 \\
    MG-CNN \cite{wang2015multiple} & VGG-19 & 83.0 & - \\
    B-CNN \cite{lin2015bilinear} & 2$\times$VGG-19&  84.8 & 90.6 \\
    FCAN \cite{liu2016fully} & ResNet-50 & 84.3 & 91.3 \\ \hline
    RA-CNN \cite{fu2017look} & 3$\times$VGG-19 & 85.3 & 92.5 \\
    MA-CNN \cite{zheng2017learning} & 3$\times$VGG-19 & 86.5 & 92.8 \\
    DT-RAM \cite{li2017dynamic} & ResNet-50 & 86.0 & 93.1 \\
    DFL-CNN \cite{wang2018learning} & ResNet-50 & 87.4 & 93.1 \\
    MAMC \cite{sun2018multi} & ResNet-50 & 86.5 & 93.0 \\
    NTSN \cite{yang2018learning} & 3$\times$ResNet-50 & 87.5 & 91.4 \\
    \multirow{2}{*}{DCL \cite{chen2019destruction}} & VGG-16 & 86.9 & 94.1 \\
    & ResNet-50 & 87.8 & 94.5 \\
    \multirow{2}{*}{TASN \cite{zheng2019looking}} & VGG-19 & 86.1 & 93.2 \\
     & ResNet-50 & 87.9 & 93.8 \\
    \multirow{2}{*}{ACNet \cite{ji2020attention}} & VGG-16 & 87.8 & 94.3 \\
    & ResNet-50 & 88.1 & 94.6 \\
    LIO \cite{zhou2020look} & ResNet-50 & 88.0 & 94.5 \\ \hline
    Ours  & ResNet-50 & \bf{88.4} & \bf{94.7} \\ \hline
    \end{tabular}
    \vspace{6pt}
    \caption{Comparison with the state-of-the-art fine-grained recognition methods on CUB-Birds \cite{wah2011caltech} and Stanford-Cars \cite{krause20133d}.}
    \label{tab:4}\vspace{-10pt}
\end{table}
\subsection{Comparison to Other Methods}\label{sec:43}

\paragraph{Fine-grained image recognition.}\label{sec:431}

In the following, we evaluate our method with state-of-the-art methods on fine-grained image recognition benchmarks including CUB-Birds \cite{wah2011caltech} and Stanford-Cars \cite{krause20133d}.
We compare with the methods using object part annotations, such as PN-CNN \cite{branson2014bird} and SPDA-CNN \cite{zhang2016spda}, using bounding box annotations, such as PA-CNN \cite{krause2015fine}, MG-CNN \cite{wang2015multiple}, B-CNN \cite{lin2015bilinear} and FCAN \cite{liu2016fully} and using only images, such as RA-CNN \cite{fu2017look}, MA-CNN \cite{zheng2017learning}, DT-RAM \cite{li2017dynamic}, DFL-CNN \cite{wang2018learning}, MAMC \cite{sun2018multi}, NTSN \cite{yang2018learning}, DCL \cite{chen2019destruction}, TASN \cite{chen2019destruction}, ACNet \cite{ji2020attention} and LIO \cite{zhou2020look}.
The results are provided in \tabref{tab:4} with a description of the backbone network, where our method achieves competitive performance.
It is noticeable that mask annotations or segmentation results from off-the-shelf algorithms \cite{he2017mask} allow us to model with 3D geometric variations while others fail even with part annotations.

\vspace{-10pt}
\paragraph{Vehicle re-identification.}\label{sec:432}
\begin{figure*}[!t]
\centering
{\includegraphics[width=1\linewidth]{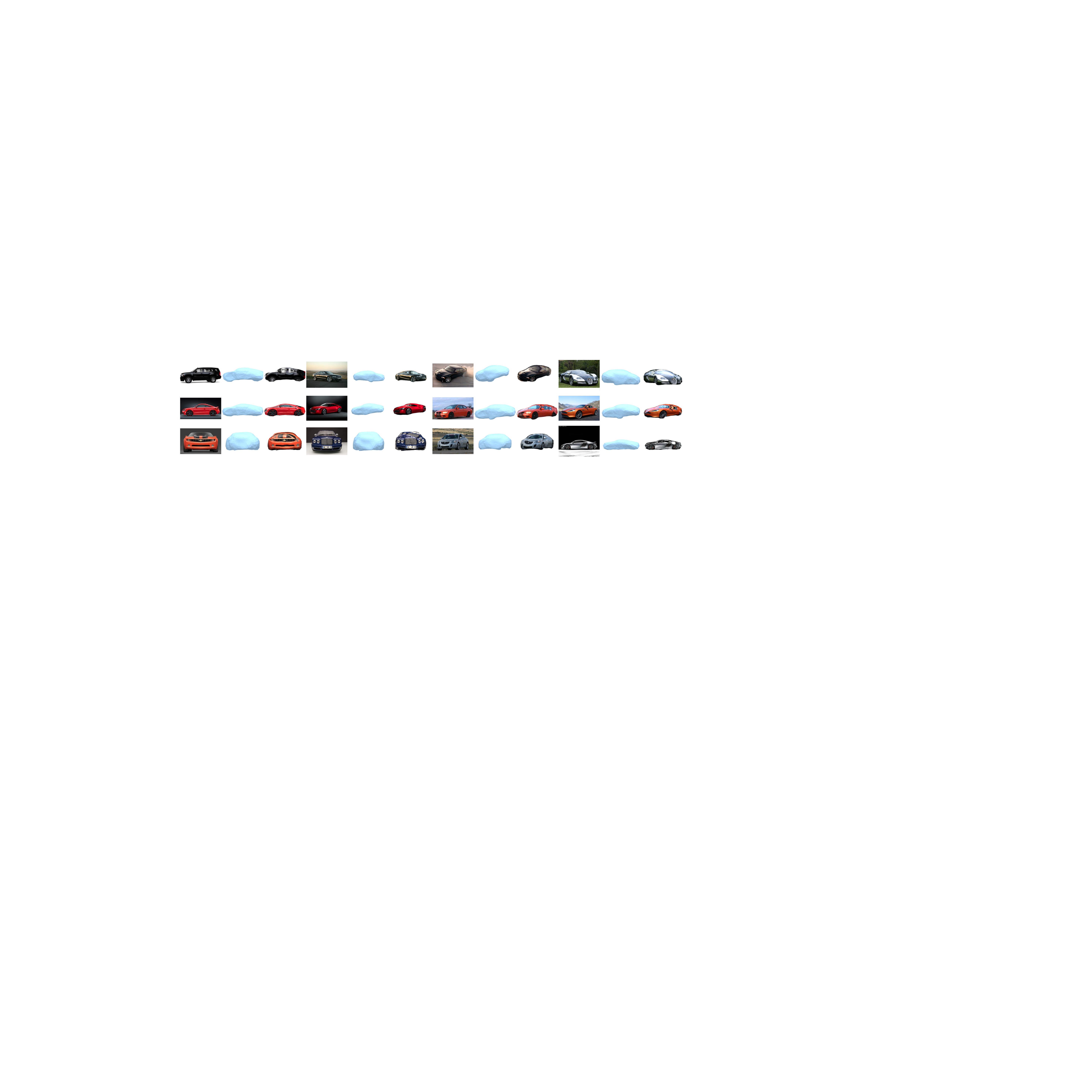}}\\
\vspace{-2pt}
\caption{\textbf{Qualitative examples on Stanford-Cars \cite{krause20133d}:} (from left) input image $I$, its 3D shape $V$, and its rendered result ${\tilde{I}}_{m}$ from predicted camera viewpoint.}
\label{fig:6}
\vspace{-10pt}
\end{figure*}

We also evaluate our method on the task of vehicle re-identification using Veri-776 benchmark \cite{liu2016deep}, where addressing subtle object variations from 2D images under different camera viewpoints is the main challenge.
Following standard practice, we use the mean average precision (mAP), Cumulative Match Curve (CMC) for top 1 (CMC@1) and top 5 (CMC@5) matches for quantitative evaluation.
We only use visual information without using either spatio-temporal information or license plate.
We compare with the state-of-the-art methods, such as OIFE \cite{wang2017orientation}, VAMI \cite{zhou2018view}, RAM \cite{liu2018ram}, PRN \cite{he2019part}, AAVER \cite{khorramshahi2019dual}, PVEN \cite{meng2020parsing} and SPAN \cite{chen2020orientation}.
As in \tabref{tab:5}, our method outperforms the conventional methods, thanks to its invariant nature under 3D geometric variation.

\vspace{-10pt}
\paragraph{3D shape reconstruction.}\label{sec:433}
\begin{table}[!t]
    \centering
    \begin{tabular}{cccc}
    \hline
    Methods &  mAP & CMC@1 & CMC@5 \\ \hline \hline
    OIFE \cite{wang2017orientation} & 0.480 & 0.659 & 0.877 \\
    VAMI \cite{zhou2018view} & 0.501 & 0.770 & 0.908 \\
    RAM \cite{liu2018ram} & 0.615 & 0.886 & 0.940 \\
    PRN \cite{he2019part} & 0.743 & 0.943 & 0.989 \\
    AAVER \cite{khorramshahi2019dual} & 0.612 & 0.890 & 0.947 \\
    PVEN \cite{meng2020parsing} & 0.795 & 0.956  & 0.984 \\
    SPAN \cite{chen2020orientation} & 0.689 & 0.940  & 0.976 \\ \hline
    Ours  & \bf{0.801} & \bf{0.959} & \bf{0.991} \\ \hline
    \end{tabular}
    \vspace{6pt}
    \caption{Comparison with the state-of-the-art vehicle re-identification methods on Veri-776 \cite{liu2016deep}.}\label{tab:5}\vspace{-10pt}
\end{table}

Finally, we compare our method with 3D reconstruction methods including CMR \cite{kanazawa2018learning}, U-CMR \cite{goel2020shape} and UMR \cite{li2020self} on CUB-Birds \cite{wah2011caltech}.
Due to the lack of ground truth 3D shape, we evaluate mask reprojection accuracy using intersection over union (IoU) as in \cite{kanazawa2018learning}.
We also measure the keypoint reprojection accuracy using a percentage of correct keypoints (PCK) with a distance threshold $\alpha=0.1$, by mapping a set of keypoints from the source image to the learned template and then to the target image via estimated shape deformation and camera viewpoint.
Note that Stanford-Cars \cite{krause20133d} and Veri-776 \cite{liu2016deep} only contain bounding box as annotations, which limit quantitative evaluation of 3D shape reconstruction.

We evaluate our method with ablation, without positional encoding and shape encoder, denoted by Ours w/o PE, ${f}_{s}$ and without shape encoder, denoted by Ours w/o ${f}_{s}$, compared to the state-of-the-art methods \cite{kanazawa2018learning,goel2020shape,li2020self} in \tabref{tab:6}.
We account that semantic keypoint \cite{kanazawa2018learning} or co-part segmentation \cite{li2020self} can be seen as a coarse-level spatial representation of the object's part (\eg, head or tail), while our method enables a dense spatial representation of the object's surface by utilizing a positional encoding module.
In addition, our shape encoder allows us to learn accurate shape deformation by discriminating shape variations in different fine-grained categories.
Since we aimed to use appearance and shape features for recognition while eliminating camera viewpoint variation, there was no significant performance gain or drop for viewpoint estimation compared to the baseline \cite{goel2020shape}.
Qualitative results on CUB-Birds \cite{wah2011caltech} and Stanford-Cars \cite{krause20133d} are in \figref{fig:5} and \figref{fig:6}.

%-------------------------------------------------------------------------
\subsection{Discussion}\label{sec:44}
Although promising results have been achieved in various tasks, several limitations still remain.
In particular, our approach is not applicable to the categories where 3D shape across instances differ significantly or undergo large articulation, \eg, human.
It is also challenging to learn 3D shape from images with heavy occlusion, which limits its applicability to generic object recognition tasks in the wild.
Nevertheless, this is an encouraging step towards understanding the underlying object variation in 3D space for image recognition tasks, and we hope it encourages future work by overcoming the aforementioned limitations.

%-------------------------------------------------------------------------

\section{Conclusion}\label{sec:5}
\begin{table}[!t]
    \centering
    \begin{tabular}{ccc}
    \hline
    \multirow{2}{*}{Methods} & \multicolumn{2}{c}{Metric} \\ \cline{2-3}
      & Mask IoU & PCK \\ \hline \hline
    CMR \cite{kanazawa2018learning} & 0.706 & 47.3 \\ 
    U-CMR \cite{goel2020shape} & 0.712  & 49.4 \\
    UMR \cite{li2020self} & 0.734 & 51.2 \\ \hline 
    Ours w/o PE, ${f}_{s}$ & 0.729 & 51.5 \\
    Ours w/o ${f}_{s}$ & 0.732 & 51.9 \\
    Ours & \bf{0.737} & \bf{52.1} \\ \hline
    \end{tabular}
    \vspace{6pt}
    \caption{Comparison of 3D mesh reconstruction on CUB-Birds dataset \cite{wah2011caltech}. Mask IoU and keypoint transfer (KT) are evaluated.}\label{tab:6}\vspace{-10pt}
\end{table}

We have introduced a novel framework to learn the discriminative representation of an object under 3D geometric variations, which accounts for the fact that object variation can be decomposed into the variation of 3D object shape and appearance in a canonical space, as well as camera viewpoint variation.
We have developed a framework for disentangling of 3D shape, appearance, and camera viewpoint variation, trained without ground-truth 3D annotation.
It allows for reconfiguring the appearance feature into a canonical space, enabling us to utilize a positional encoding for better representation learning.
To deal with subtle object variation and improve recognition ability, we have further introduced a shape encoder to utilize 3D shape deformation.
Our experiments have shown that our method effectively learns the discriminative object representation on fine-grained recognition and 3D shape reconstruction tasks.

%-------------------------------------------------------------------------

{\small
\bibliographystyle{ieee_fullname}
\bibliography{egbib}
}

\end{document}